%% file: main.tex
\documentclass[conference]{IEEEtran}
\IEEEoverridecommandlockouts

\input{bangla_commands}
\usepackage{courier}
\usepackage[T1]{fontenc}
\usepackage{times}

\usepackage{mathptmx}
\usepackage{amsmath,amssymb,amsfonts}
\usepackage{algorithmic}
\usepackage{microtype}
\usepackage{graphicx}
\usepackage[table]{xcolor}
\usepackage{ragged2e}
\usepackage{tabularx}
\usepackage{siunitx}
\usepackage{booktabs}  
\usepackage{textcomp}
\usepackage{xcolor}
\usepackage{array}
\usepackage[sorting=none]{biblatex}
\usepackage{fancyhdr}

\def\BibTeX{{\rm B\kern-.05em{\sc i\kern-.025em b}\kern-.08em
    T\kern-.1667em\lower.7ex\hbox{E}\kern-.125emX}}

\begin{document}

\title{Enhancing Bangla Language Next Word Prediction and Sentence Completion through Extended RNN with Bi-LSTM Model On N-gram Language\\

}

\author{
\IEEEauthorblockN{Md Robiul Islam\textsuperscript{1}, Al Amin\textsuperscript{2}, Aniqua Nusrat Zereen\textsuperscript{3}}
\IEEEauthorblockA{
{\textsuperscript{1,2} Department of Computer Science and Engineering, Uttara University, Dhaka, Bangladesh}} 
{\textsuperscript{3} Department of Computer Science and Engineering, BRAC University, Dhaka, Bangladesh}\\
\{robiul.cse.uu, alaminbhuyan321, aniqua.zereen\}@gmail.com}

\maketitle

\begin{abstract}
Texting stands out as the most prominent form of communication worldwide. Individual spend significant amount of time writing whole texts to send emails or write something on social media, which is time consuming in this modern era. Word prediction and sentence completion will be suitable and appropriate in the Bangla language to make textual information easier and more convenient.
This paper expands the scope of Bangla language processing by introducing a Bi-LSTM model that effectively handles Bangla next-word prediction and Bangla sentence generation, demonstrating its versatility and potential impact. We proposed a new Bi-LSTM model to predict a following word and complete a sentence. We constructed a corpus dataset from various news portals, including bdnews24, BBC News Bangla, and Prothom Alo. The proposed approach achieved superior results in word prediction, reaching 99\% accuracy for both 4-gram and 5-gram word predictions. Moreover, it demonstrated significant improvement over existing methods, achieving 35\%, 75\%, and 95\% accuracy for uni-gram, bi-gram, and tri-gram word prediction, respectively. 
\end{abstract}

\begin{IEEEkeywords}
Bangla word prediction, Bangla sentence completion, LSTM, Bangla language.
\end{IEEEkeywords}

\section{Introduction}\label{introduction}
In the modern era, the ubiquity of communication devices is indisputable. However, their intrinsic dependence on laborious and time-intensive text input procedures constitutes a noteworthy drawback. In response to these limitations, integrating textual word prediction systems has emerged as a promising remedy. Such systems mitigate the challenges associated with repetitive typing, by effectively streamlining the overall text input process. By utilizing the predictive abilities embedded within text prediction systems, the need for manual input of text is eliminated. As an alternative, the most popular word suggestions are shown to the user, maximizing both user enjoyment and efficiency. This advanced forecasting procedure creates a smooth and user-friendly text entry interface, significantly improving the speed and fluidity of digital communication \cite{one}. This clever application significantly lowers the number of keystrokes needed for word entry, which helps to mitigate writing obstacles. The effectiveness of the result is directly correlated with the number of keystrokes that the user saves, since this lowers the amount of time invested as well as the effort involved in producing text \cite{two}. Once a word or phrase is entered, the program automatically creates a curated list of possible lexical possibilities. When the user finds the word they want in this list, selecting it is a simple process that only requires one click to insert the selected word into the page \cite{three}. Scholars have conducted numerous investigations to recognize the benefits that come with word prediction systems, using a range of approaches to improve word prediction algorithms and enhance user experience in general. Specifically, within the parameters of their research project, they have diligently investigated and applied various methods to enhance the effectiveness and user-friendliness of word prediction systems. Partha et. al. utilized a combination of Recurrent Neural Network (RNN) and Long Short-Term Memory (LSTM) to predict the next word in the Assamese language. For untranslated Assamese text, they obtained an accuracy of 88.20\%, whereas for phonetically transcribed Assamese language, they acquired a 72.10\% accuracy \cite{four}. The model presented by the Steffen Bickel et. al. accurately predicts the most likely word that will be typed next \cite{five}.  They used weather forecasts, food recipes, call center communications, and personal emails to assess their model. They modified N-gram language models to predict the following words to evaluate the model's performance, placing more emphasis on precise prediction than conventional performance indicators. With a focus on the English language, our study has investigated various word prediction methods for other languages. Nevertheless, few thorough investigations have been conducted with the goal of creating word prediction models for the Bangla language or sentence prediction \cite{six}. The model's efficacy is intricately tied to the quality and diversity of the dataset utilized during the training phase. To this end, a comprehensive dataset has been meticulously curated from diverse sources, including bdnews24 \cite{seven}, Prothom Alo \cite{eight}, and BBC Bangla news \cite{nine}. Despite the considerable research on enhancing word prediction capabilities within the Bengali language using machine learning methodologies, persistent advancements are imperative to augment system efficiency and precision. The ongoing evolution of these systems hinges upon their capacity to adeptly handle expansive datasets, thereby facilitating more nuanced and accurate prognostications.

The comprehensive allowance of this research work is -
\begin{itemize}
   \item No previous research has utilized this technique for Bangla language word prediction.
   \item Utilizing a large dataset for Bangla word prediction, significantly improves the accuracy of our suggested approach.
   \item The effectiveness of our suggested approach exceeds that of other approaches employed for Bangla word prediction and sentence completion.
\end{itemize}

The remaining sections of this research are structured like this: part \ref{relatedWork} discusses the previous researcher’s work, part \ref{methodology} outlines the methodology encompassing dataset creation and implementation with pre-processing, part \ref{resultEvaluation} presents the experimental result and part \ref{conclusion} discussed the constraint and upcoming directions.

\section{Related Works}\label{relatedWork}

Predicting the next word is a big focus in Natural Language Processing (NLP) research. In this context, discerning the following word presents a multifaceted challenge. A notable obstacle arises from the inherent nature of machines, which predominantly interpret information through ASCII values, essentially binary code represented as 0s and 1s. The need to represent linguistic elements in numerical form becomes imperative. Researchers have diligently addressed these challenges associated with the next-word prediction. Their efforts have been directed towards finding effective solutions to the intricacies posed by machine comprehension limited to binary representations. They aim to transform words into numerical entities, paving the way for more accurate and sophisticated next-word predictions in natural language scenarios. The majority of the work has been LSTM based. To tackle predicting the following word in sequential information, Mikolov et al. suggested a language model using RNN \cite{ten}, tasks that were previously challenging for RNN to solve were solved by Felix et al. using LSTM \cite{eleven}. Sutskever et al. used an RNN model to predict the next word in a sequence of data. Their recently presented model was a development of RNNSearch \cite{twelve}, and it was trained from a question-answering model to a language model via backpropagation. Zhou et al. proposed the C-LSTM model a while back, which combines RNN with LSTM for classification and prediction on phrase representations \cite{thirteen}. A method for creating Bengali text using a sequence paradigm was presented by Banik et al. \cite{fourteen}. Another approach was presented by Abujar et al. \cite{fifteen}, who employed bi-directional RNN to generate Bengali text In their study, Barman et al. employed an RNN-based methodology to develop a next-word prediction model \cite{four}. In essence, they created a model of the LSTM, a unique kind of RNN. They used their work to suggest the subsequent word with the highest likelihood reliably next. The N-gram has been used to reduce the prediction time while typing in the Kurdish language, and the model was successful in prediction. This model has been developed in the R programming language. The maximum accuracy recorded in this model is 96.3\% \cite{sixteen}.

\section{Methodology}\label{methodology}

\begin{figure}[!ht]
    \centering
    \includegraphics[scale=0.6]{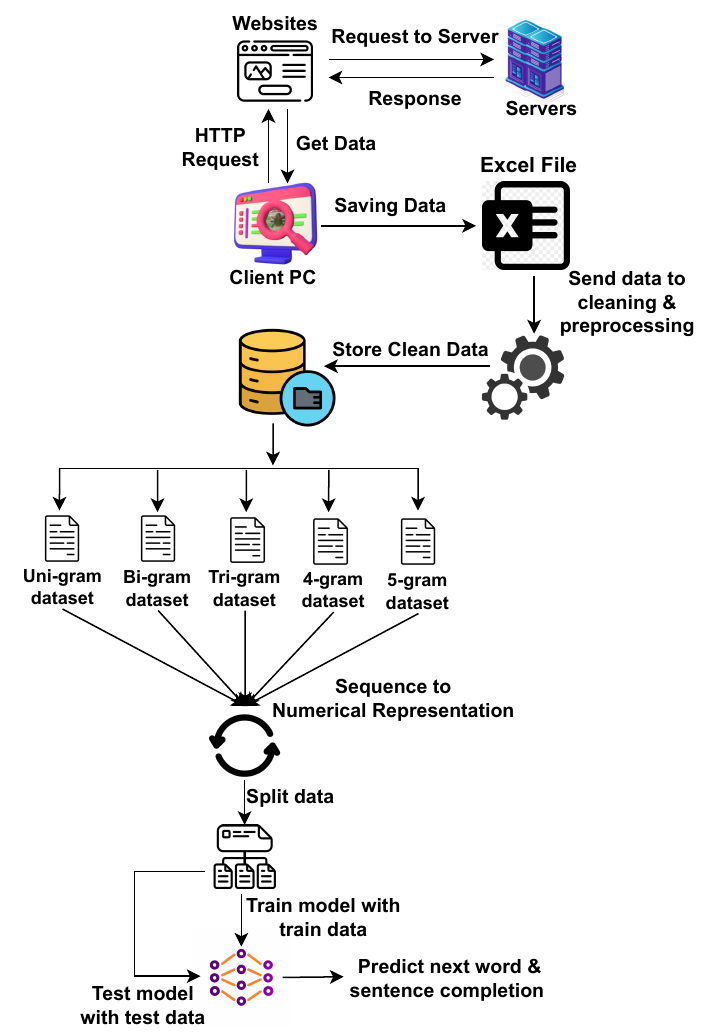}
    \caption{\centering Proposed methodology}
    \label{fig:methodology}
\end{figure}

This section describes the proposed methodology as well as the operating principles of the machine learning algorithms. The Bi-LSTM model predicted the next word and sentence in Bangla. Fig. \ref{fig:methodology}. illustrates the full process of the methodology.

We began by introducing the dataset in subsection A, followed by data preprocessing techniques in subsection B, and culminated in the model implementation details in subsection C.

\subsection{Dataset Outline}
To evaluate the proposed approach, we assembled a large corpus of Bangla text, totaling 1.7 GB, gathered from diverse sources. Table \ref{table:dataSummary} provides a statistical overview of the collected data.

\begin{table}[!htt]
  \centering
  \caption{\centering Data Summary}
  \label{table:dataSummary}
  \begin{tabular}{ccc}
    \toprule
    \textbf{Data Source} & \textbf{Distinct words} & \textbf{Total} \\
    \midrule
       Bdnews24 \cite{seven} &	45789 &	3892651\\
       Prothom Alo \cite{eight}	& 97000 &	2087916 \\
       BBC News Bangla \cite{nine} & 	67004 & 4045683\\
       
    \bottomrule
\end{tabular}
\end{table}

\subsection{Data Preprocessing}
\begin{figure}[!ht]
    \centering
    \includegraphics[scale=0.4]{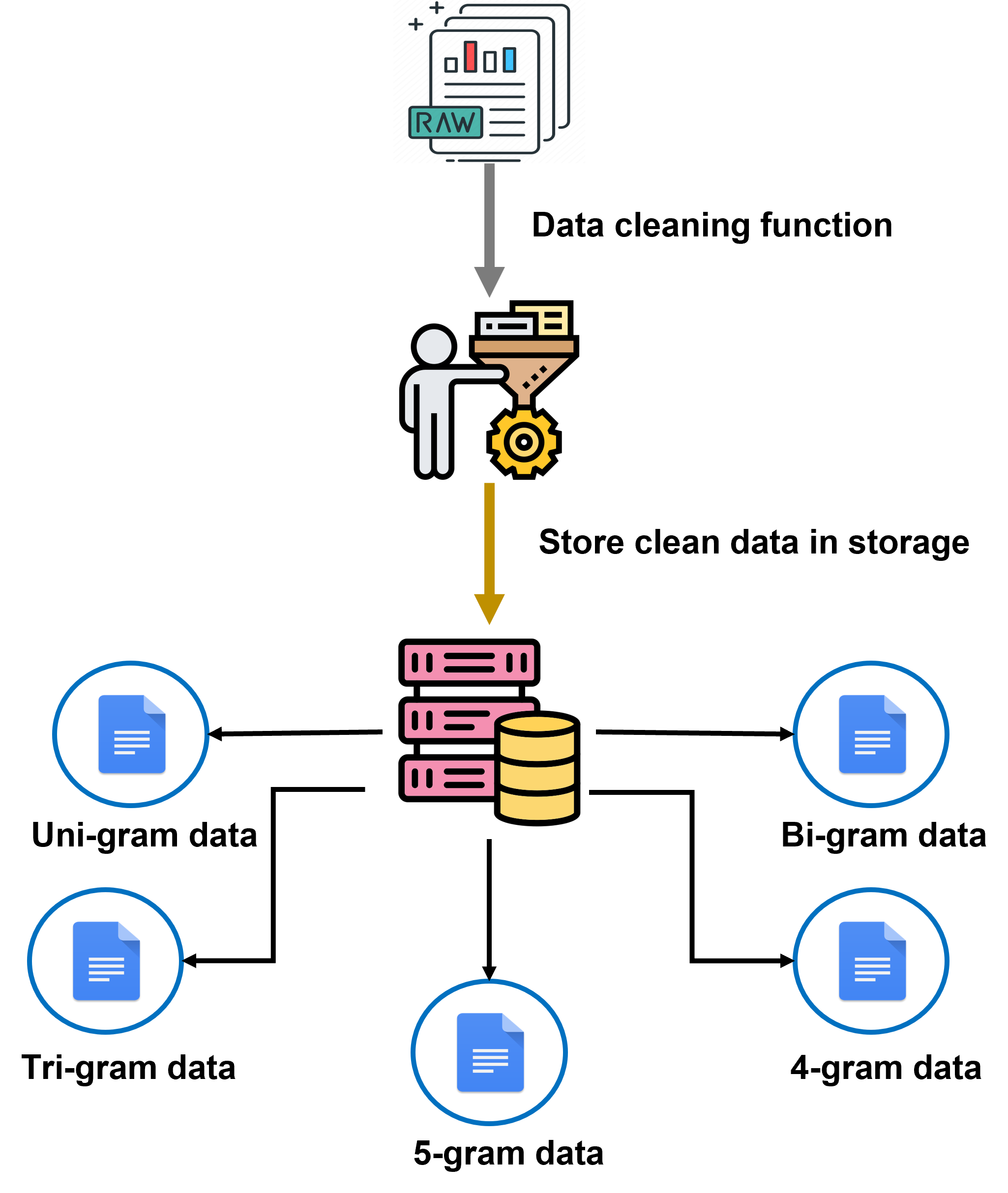}
    \caption{\centering Data Preprocessing Workflow}
    \label{fig:dataprocessing}
\end{figure}

After collecting the data, we used several functions to remove unwanted commas, numbers, etc. from our dataset and five distinct datasets were generated from the cleaned dataset: Unigram, Bigram, Trigram, 4-gram, and 5-gram. Fig. \ref{fig:dataprocessing}. effectively illustrates the data-cleaning workflow and the creation of different n-gram datasets. Language models, known as N-grams, are statistical models that calculate the likelihood of a specific word sequence occurring in a language. In this study, we have used n-gram language models to capture the sequential relationships found in Bangla text \cite{eighteen}. When attempting to predict the next word or words, the quantity of input words can typically vary from one attempt to the next. One word, a 1-gram or uni-gram, is produced whenever there is only one input word \cite{nineteen}. Bi-gram is used when the input consists of two words, and the output is one word \cite{nineteen}. Furthermore, a Three-gram, sometimes known as a Tri-gram, is created when three words are input and one word is output \cite{nineteen}, and analogously for Four-gram and Five-gram. Usually, the last four or five words are enough to grasp the sequence's dependency. To better understand n-gram language models, consider an example of a Bangla sentence. Analogous to uni-grams and bi-grams, tri-grams, and further n-grams maintain the pattern of taking multiple input values and generating a single output value, demonstrating a consistent approach to sequence processing.

\begin{table}[ht]
  \centering
  \caption{\centering Bi-gram Model Data Sample}
  \label{table:bigram}
  \begin{tabular}{ccc}
    \toprule
    \textbf{Input\_1} & \textbf{Input\_2} & \textbf{Output} \\
    \midrule
       {\bng baNNGlaedsh } & {\bng EkiT } & {\bng smrRd/dh } \\
       {\bng EkiT } & {\bng smrRd/dh } & {\bng sNNGs/krRitr } \\
       {\bng smrRd/dh } & {\bng sNNGs/krRitr } & {\bng edsh } \\

    \bottomrule
\end{tabular}
\end{table}

\begin{table}[!ht]
  \centering
  \caption{\centering Tri-gram Model Data Sample}
  \label{table:trigram}
  \begin{tabular}{cccc}
    \toprule
    \textbf{Input\_1} & \textbf{Input\_2} & \textbf{Input\_3}& \textbf{Output} \\
    \midrule
       {\bng baNNGlaedsh } & {\bng EkiT } & {\bng smrRd/dh } & {\bng sNNGs/krRitr } \\
       {\bng EkiT }  & {\bng smrRd/dh } & {\bng sNNGs/krRitr } & {\bng edsh } \\

    \bottomrule
\end{tabular}
\end{table}

Table \ref{table:bigram} and Table \ref{table:trigram} showing the Bi-gram and Tri-gram models. It shows how it predicts the next word based on the inputted word. For Bi-gram, only two words are used to predict the next word; in the case of Tri-gram, there are three words used to predict the next word.

\subsection{Implementation}
The n-gram language model determines the likelihood of each potential next word and chooses the one with the greatest likelihood to be the next word or words. The utilization of a probabilistic viewpoint enables us to determine the probability of the next value, empowering us to predict the most probable next value based on the input value {\bng ekRmilen eDRan }, equation (1) has been used to get the highest likelihood of the next word.


\[
\begin{split}
    \text{Word after that} &= \textbf{Max} \bigg( P(\text{{\bng Hamlar}} | \text{{\bng ekRmilen eDRan}}), \\
    &\phantom{=} P(\text{{\bng ghTnaiT}} | \text{{\bng ekRmilen eDRan}}), \\
    &\phantom{=} P(\text{{\bng Aasel}} | \text{{\bng ekRmilen eDRan}}) \bigg)
\end{split}
\]


Due to the frequent emergence of zero probability, the model cannot recommend the next word with the highest probability, leading to the failure of the entire method in predicting the most suitable value. The effectiveness of n-gram models diminishes when confronted with large datasets featuring lengthy input sequences or a considerable number of n-grams. To alleviate the impact of data sparsity, Back-off, and Katz Back-off techniques were employed to refine the probability distribution, enabling the effective management of n-grams with low counts \cite{tweenty}. RNNs keep updating their memory with past information to maintain a consistent understanding.


Long sequences cause RNNs to lose the impact of previous layers, which gives rise to the vanishing gradient problem. Two well-known gating mechanisms, LSTM (Long Short Term Memory) and GRU (Gated Recurrent Unit) were used in response to the vanishing gradient problem in RNN. These mechanisms allow the network to efficiently capture long-term dependencies in sequential data. The use of three gating mechanisms—input, forget, and output—by Bi-LSTM in this study justifies its adoption since they improve the model's capacity to learn from and retain information from distant past inputs, hence resolving the drawbacks associated with vanishing gradients in RNN. GRU employs update and reset gates to handle sequential data, whereas LSTM's ability to maintain internal memory over time makes it more effective for processing complex datasets compared to GRU, which doesn't have a specialized forget gate.

\begin{figure}[!ht]
    \centering
    \includegraphics[scale=0.37]{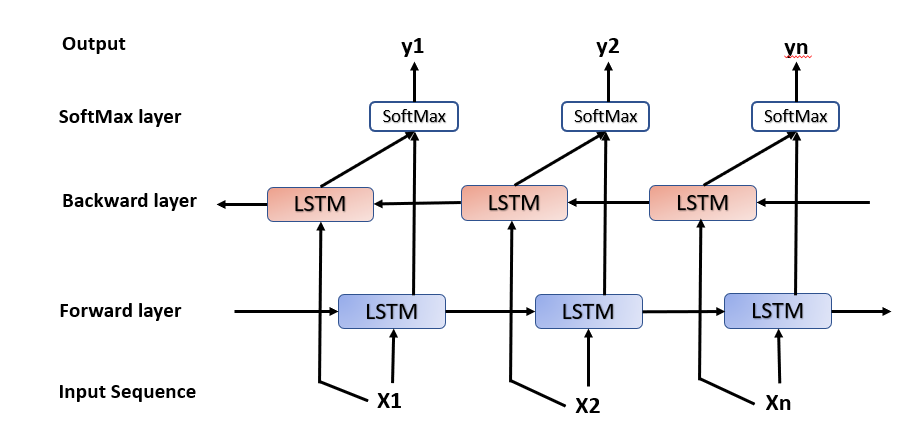}
    \caption{\centering Structure of Bi-LSTM recurrent neural networks}
    \label{fig:biLstmRnnLayers}
\end{figure}

Two separate recurrent neural networks (RNNs) are integrated to create bidirectional RNNs. This setup lets the network look at information from both directions (like before and after) for each point in the sequence. When employing bidirectional processing, the input data is traversed in two directions: from past to future and from future to past. The key distinction from unidirectional RNNs lies in using a bidirectional long short-term memory (Bi-LSTM) structure, where the backward LSTM retains information from the future. By combining the hidden states from the forward and backward LSTMs, the model can capture contextual information from the sequence's preceding and succeeding elements. This bidirectional approach, exemplified by Bi-LSTMs, yields notable performance improvements as it enhances the network's understanding of contextual nuances. Fig. \ref{fig:biLstmRnnLayers} showing the Bi-LSTM layer architecture.


\begin{figure}[!ht]
    \centering
    \includegraphics[scale=0.6]{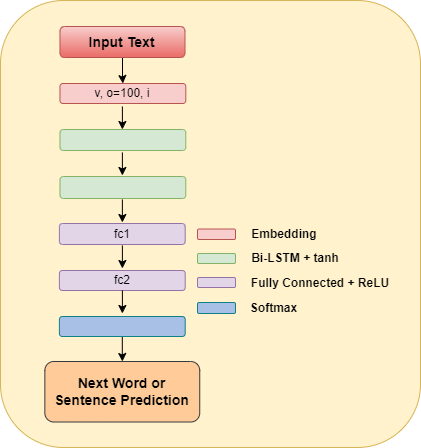}
    \caption{\centering The proposed Bi-LSTM model architecture}
    \label{fig:modelSummary}
\end{figure}


Fig. \ref{fig:modelSummary}. shows the architecture of the trained models with five hidden layers named Embedding, Bi-LSTM Layer1, Bi-LSTM Layer2, Fully Connected Layer1, Fully Connected Layer2. For Bi-LSTM layers we used 100 LSTM units with \textbf{\textit{tanh}} activation function and for Fully Connected Layer1 we used \textbf{\textit{ReLU}} and for Fully Connected Layer2 we used \textbf{\textit{softmax}} activation function. In the Embedding layer \textbf{\textit{v}} denotes input dimension or vocabulary length, \textbf{\textit{o}} represent the output dimension and \textbf{\textit{i}} represent the input length.

\subsection{Word Prediciton}
Upon effectively completing training on all five datasets (U-gram to 5-gram), we have obtained five unique trained models optimized for processing input of varying lengths.

\begin{figure}[!ht]
    \centering
    \includegraphics[scale=0.5]{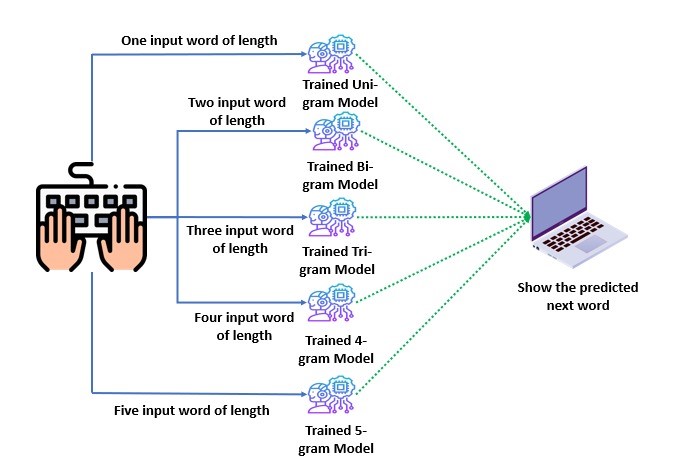}
    \caption{\centering Next word prediction from the model}
    \label{fig:wordPrediction}
\end{figure}

These models take in sequences of words of different lengths and guess what word is likely to come next after the input sequence. When the input sequence contains only one word, the Uni-gram model is utilized for training. With a single-word input as its input, the model will most likely anticipate the next word. Systematically, In the case of two input words, the inputted sequence is directed to the trained Bi-gram model, designed to handle two-word inputs and predict an output word. Similarly, for the other trained models. Fig. \ref{fig:wordPrediction}. demonstrates the word prediction methodology for input sequences of different lengths using five distinct trained models. In cases where the input word sequence exceeds five words, the 5-gram model relies solely on the last five words to predict the next word. The last five words typically provide enough context to determine the next word.

\subsection{Sentence Prediction}
\begin{figure}[!ht]
    \centering
    \includegraphics[scale=0.8]{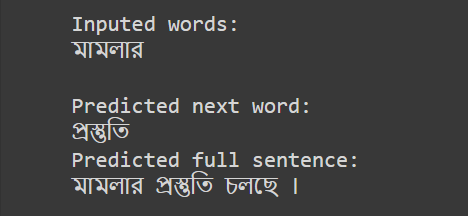}
    \caption{\centering Next word prediction and sentence completion}
    \label{fig:sentPrediction}
\end{figure}

Our innovative model combines N-gram training with Bi-LSTM-based RNNs. It iteratively predicts and appends the next word to the input sequence, capturing evolving context for accurate predictions. The process ends upon encountering sentence-ending punctuation. In Bangla, sentence termination is signified by the symbols "|" for declarative statements and "?" for interrogative statements. Once the model detects an end punctuation mark, it ceases prediction and outputs the complete sentence, representing a cohesive and meaningful continuation of the given word sequence. Fig. \ref{fig:sentPrediction}. shows the word and sentence prediction by the proposed model.

\section{Result Evaluation}\label{resultEvaluation}
\begin{figure}[!ht]
    \centering
    \includegraphics[scale=0.5]{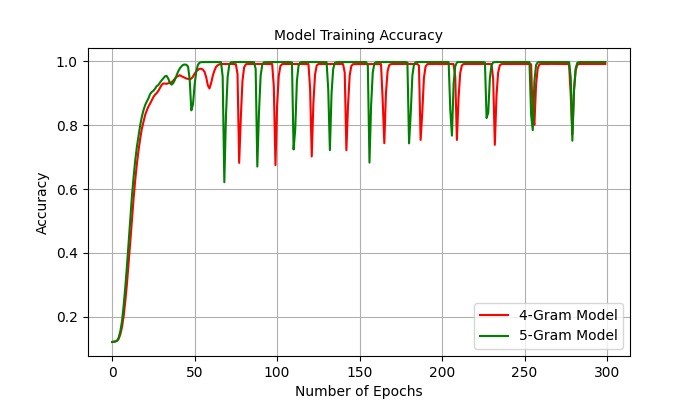}
    \caption{\centering 4-Gram and 5-Gram model accuracy per epochs}
    \label{fig:acc}
\end{figure}

\begin{figure}[!ht]
    \centering
    \includegraphics[scale=0.5]{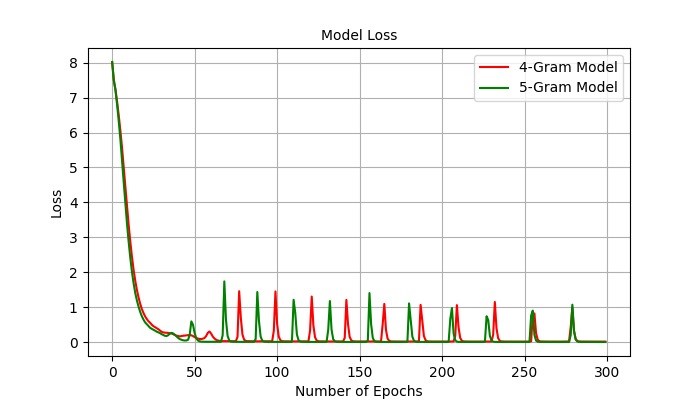}
    \caption{\centering 4-Gram and 5-Gram model loss per epochs}
    \label{fig:loss}
\end{figure}

We have evaluated our proposed approach using a corpus dataset, training five different models with identical configurations for up to 300 epochs. Fig. \ref{fig:acc}. depicts the accuracy across epochs, while Fig. \ref{fig:loss}. showcases the loss over the same epochs. Our 4-gram and 5-gram models exhibit an average accuracy of 99\% and 99.74\%, respectively, accompanied by average losses of 2.04\% and 1.11\%. 


\begin{table}[h]
\caption{Comparative Analysis}
\label{table:comparisonTable}
\centering
\begin{tabular}{|c|c|}
\hline
\textbf{Reference}  & \textbf{Accuracy} \\ [1 ex]
\hline
Barman et. al. \cite{four} & 88.20\% \\ [1 ex]
Hamarashid et. al. \cite{sixteen} & 96.30\% \\ [1 ex]
\textbf{Proposed Model} &  \textbf{99.00\%} \\ [1 ex]
\hline
\end{tabular}
\end{table}

Table \ref{table:comparisonTable} shows the comparison between Barman et. al. \cite{four} and Hamarashid et. al. \cite{sixteen} and our proposed system. In Author \cite{sixteen}, they used Tri and a Hybrid Approach of Sequential LSTM and N-gram; their accuracy was 88.20\%. On the other hand, we have a maximum efficiency of 99\% for high-order sequences.

\section{Conclusion}\label{conclusion}
The word prediction and sentence completion model shows promising accuracy based on the provided dataset. Using multiple pattern-discovery techniques is essential for effectively removing noisy data in NLP. Next-word prediction is an example of an NLP application because it involves text mining. The Bi-LSTM model was used for forecasting over multiple epochs, resulting in a significant decrease in loss.
Specific preprocessing steps and model modifications could be implemented to improve the model's prediction performance. We will make our data set public so that researchers can conduct research in the near future. A more diverse dataset will improve our work.

\printbibliography
\end{document}

%% file: bangla_commands.tex
\def\bng{\bngx}

\font\bngx=bang10

\def\*#1*#2{o\null{#2}{#1}}

\def\sh#1{\setbox0=\hbox{#1}%
     \kern-.02em\copy0\kern-\wd0
     \kern.04em\copy0\kern-\wd0
     \kern-.02em\raise.0433em\box0 }